\documentclass{article}

\usepackage[sglblindworkshop, final]{neurips_2026}
\makeatletter
\renewcommand{\@noticestring}{Accepted at the ACM CAIS 2026 Workshop on AI Agents for Discovery in the Wild.}
\makeatother

\usepackage[utf8]{inputenc}
\usepackage[T1]{fontenc}
\usepackage[hypertexnames=false]{hyperref}
\usepackage{url}
\usepackage{booktabs}
\usepackage{amsmath}
\usepackage{amsfonts}
\usepackage{nicefrac}
\usepackage{microtype}
\usepackage{xcolor}
\usepackage{graphicx}
\usepackage{algorithm}
\usepackage{algpseudocode}
\usepackage{tikz}
\usetikzlibrary{positioning,arrows.meta,shapes.geometric}

\title{Stage--Audit: Auditable Source-Frontier Discovery for Cross-Wiki Tables}

\workshoptitle{AI Agents for Discovery in the Wild}

\author{%
  Chen Shen \\
  Megagon Labs \\
  \texttt{chen\_s@megagon.ai}
}

\begin{document}
\maketitle

\begin{abstract}
LLM-curated tables can appear source-grounded while containing unsupported rows: the curator may recall entries from parametric memory and retroactively attach page-level citations that are not the actual source.
We study this hazard in \emph{Seed2Frontier discovery}: the task of finding complement Wikipedia pages from a seed page to assemble a structured table.
Stage--Audit addresses it with disjoint curator-auditor write rights, a row-level source-citation gate, and a 12-check audit taxonomy over keys, schema, source roles, cardinality, and scope.
On a curated 51-instance Seed2Frontier evaluation set spanning 15 top-level domains, Stage--Audit improves source-frontier precision over a vanilla LLM curator from 0.356 to 0.505 ($+42\%$ relative) and F1 from 0.334 to 0.451 ($+35\%$), while maintaining explicit per-row source traceability.
The vanilla-LLM-vs-Stage--Audit comparison isolates the policy contribution rather than LLM-based discovery in general.

\end{abstract}

\section{Introduction}

Take a question like ``list transboundary rivers longer than 1000\,km by continent.''
No single Wikipedia page carries the answer.
An LLM agent that wants to return a structured table, not a paragraph, has to walk a master list, the continent overview pages, and the individual river articles, then stitch them into rows that are keyed and source-traceable.
We call this problem \emph{Seed2Frontier discovery}: starting from a seed Wiki page, discover the source frontier needed to assemble a structured table.

The technical hazard is unintentional ungroundedness.
An LLM curator can emit rows recalled from parametric memory and retroactively attach plausible page-level citations; the row looks cited even when the citation is not the actual source of the row.
Prior work documents self-preference and weak self-correction at evaluation time \citep{panickssery2024selfrecognition,wataoka2024selfpreference,barkan2025overconfidence,huang2024cannot,kamoi2024selfcorrection}; here the failure happens earlier, when the reference table is built.

We introduce \emph{Stage--Audit}, a governance protocol for Seed2Frontier.
A curator can stage canonical rows only with an external source URL and locator; an auditor cannot edit the table and instead appends findings under a 12-check taxonomy over row evidence, primary keys, schema, source roles, cardinality, and scope (Figure~\ref{fig:system}).
Existing retrieval systems can attempt Seed2Frontier (Section~\ref{sec:related}); our contribution is the governance overlay, not a new retrieval method.
A table can have individually plausible rows and still be unusable if its partition, key, or exhaustiveness claim is wrong, which is what our table-level audit catches.

We make three contributions.
\textbf{(1) Task and curated set:} a 51-instance Seed2Frontier evaluation set spanning 15 top-level domains, with Wikipedia seed pages, labelled complement pages, and primary-key ground-truth tables.
\textbf{(2) Protocol:} Stage--Audit defines role-disjoint write rights, a source-citation gate, a row-witness property, and a 12-check audit taxonomy for structured tables.
\textbf{(3) Policy ablation:} a four-configuration comparison (memory-only, seed-outlink, vanilla LLM curator, Stage--Audit) shows that the source-citation gate plus audit, not LLM-based discovery alone, drives source-frontier precision and F1.

\section{Related Work}
\label{sec:related}

\textbf{Source-grounding and judge bias.}
FActScore, ALCE, SAFE, and PaperTrail evaluate generated text after the fact \citep{min2023factscore,alce2023,wei2024safe,martinboyle2026papertrail}; LLM-as-judge work documents bias and self-preference under same-family setups \citep{zheng2023mtbench,panickssery2024selfrecognition,wataoka2024selfpreference}.
Stage--Audit places the gate before a row enters the table, so the output is a structured table with keys and scope claims rather than a sequence of atomic facts.

\textbf{Cross-page Wiki retrieval and citation-aware generation.}
SRAG, KARMA, WikiContradict, and InfoGather bear on multi-page Wiki construction \citep{srag2025,karma2025,wikicontradict2024,infogather2012}; HybridQA, OTT-QA, and Open-WikiTable evaluate QA over fixed multi-page table corpora \citep{chen2020hybridqa,chen2021ottqa,kweon2023openwikitable}, whereas Seed2Frontier evaluates \emph{discovery} of the source frontier itself.
Our vanilla-LLM-curator baseline stands in for unguarded LLM-based retrieval of this family.
RARR retrofits citations to free-text outputs and Self-RAG learns to self-cite during generation \citep{gao2023rarr,asai2024selfrag}; both place the citation step inside the generator and have no separate audit role over a structured table contract.
Stage--Audit instead enforces citation \emph{before} a row is staged and adds a disjoint auditor whose findings cannot rewrite the curator's table; SPADE and MAST motivate the table-contract view \citep{shankar2024spade,cemri2025mast}, while generator--critic, debate, Self-Refine, and Constitutional AI explore critique-via-extra-calls without a row-level source gate \citep{madaan2023selfrefine,du2023debate,bai2022constitutional}.

\section{Method}
\label{sec:method}

\subsection{Stage--Audit Oversight Protocol}
\label{sec:method:protocol}

Let $M$ be a model or agentic framework used to curate source-grounded tables for user questions.
Some curated tables may later be used to evaluate or diagnose the same model family.
The threat is not adversarial deception but unintentional self-reference: $M$ emits remembered rows and then attaches plausible page-level citations.
Because intrinsic self-correction is weak without external feedback \citep{huang2024cannot,kamoi2024selfcorrection}, the protocol makes the source policy, not the model's self-critique, the load-bearing mechanism.

\begin{figure}[t]
  \centering
  \resizebox{0.96\linewidth}{!}{%
  \begin{tikzpicture}[
    node distance=5pt,
    box/.style={draw, rounded corners=2pt, align=left, inner sep=4pt, font=\scriptsize, text width=11.6cm},
    qbox/.style={box, fill=blue!6},
    cbox/.style={box, fill=green!7},
    abox/.style={box, fill=orange!8},
    rbox/.style={box, fill=red!6},
    obox/.style={box, fill=gray!10},
    arrow/.style={->,>=Latex,thick},
  ]
  \node[qbox] (q) {\textbf{User query.} ``List transboundary rivers $\geq 1000$\,km by continent, with length and traversed countries.''};
  \node[obox, below=of q] (s) {\textbf{Seed.} \texttt{en.wikipedia.org/wiki/Transboundary\_river}};
  \node[cbox, below=of s] (c) {\textbf{Curator (stage locator-backed rows).}\\
  Proposes 13 complement pages (e.g., \emph{Senegal\_River}, \emph{Amur}, \emph{Ural\_(river)}).\\
  Stages 31 rows; each row carries \texttt{source\_page} and a section/infobox locator.};
  \node[abox, below=of c] (a) {\textbf{Auditor (append-only findings).}\\
  \(\checkmark\)\ Source-citation gate: every row cites a page and locator.\\
  \(\triangle\) \textsc{scope, blocking}: \emph{Ural} crosses Europe/Asia; partition label ambiguous.\\
  \(\triangle\) \textsc{cardinality, blocking}: 31 emitted vs.\ 33 expected from frontier coverage.};
  \node[rbox, below=of a] (r) {\textbf{Repair (curator restages).} 2 missing rows pulled from \emph{List\_of\_transboundary\_rivers}; \emph{Ural} continent set to \texttt{Europe; Asia}.};
  \node[obox, below=of r] (o) {\textbf{Audited table.} 33 rows $\times$ 4 cols, primary key=\emph{River}, every row $\to$ Wiki locator.};
  \node[box, below=of o, fill=gray!18] (h) {\textbf{Human acceptance (deployed).}\ Reviewer signs off on audited table; \emph{ablated in our experiment} (LLM auditor only).};
  \draw[arrow] (q) -- (s); \draw[arrow] (s) -- (c); \draw[arrow] (c) -- (a); \draw[arrow] (a) -- (r); \draw[arrow] (r) -- (o); \draw[arrow,dashed] (o) -- (h);
  \draw[->,>=Latex,thin] (r.east) -- ([xshift=0.5cm]r.east) -- node[right,font=\scriptsize] {restage} ([xshift=0.5cm]c.east) -- (c.east);
  \end{tikzpicture}}
  \caption{Stage--Audit walkthrough on a curated Seed2Frontier instance. The curator proposes locator-backed rows; the auditor appends findings under the 12-check taxonomy and cannot edit the table; blocking findings trigger a repair-restage loop back to the curator. The dashed final step shows the deployed human-acceptance role; our experiment ablates the human step and runs a single curator-and-auditor pass with no repair-restage iteration, to isolate the single-step value of the policy.}
  \label{fig:system}
\end{figure}
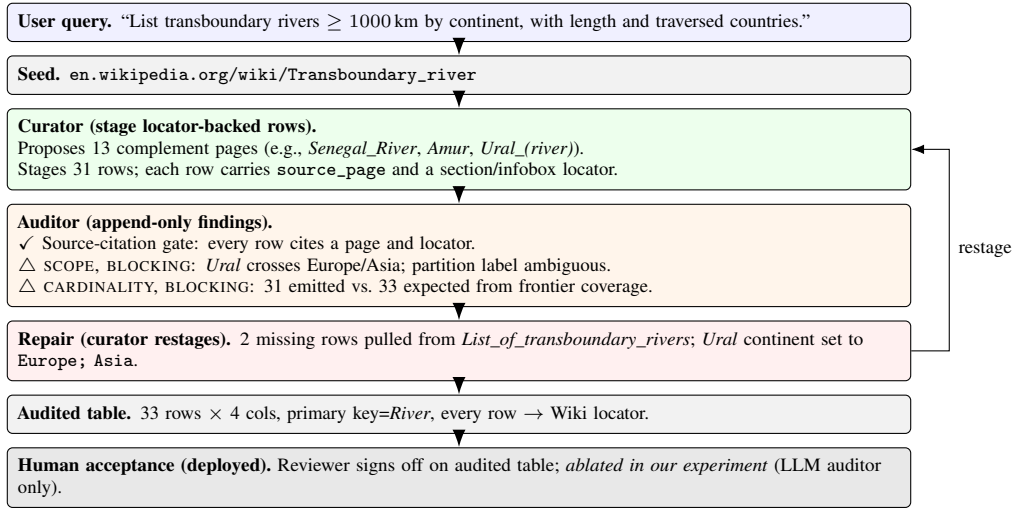

\noindent\textbf{Design tenets.}
Stage--Audit has four operational rules.
\emph{Append-only audit}: the auditor cannot edit the canonical table and records findings separately.
\emph{Cited-source-first}: every emitted row must carry a source URL and locator before staging.
\emph{Parametric memory is not evidence}: remembered facts may suggest where to search, but cannot justify a row.
\emph{Two-axis severity}: every finding is tagged by issue type (factual / structural / scope-related) and severity (blocking / hygiene / suggestion), so repair can distinguish blockers from cosmetics.

\noindent\textbf{Artifact lifecycle.}
A staged table moves through four states (\emph{proposed}, \emph{staged}, \emph{audited}, \emph{repaired}), iterating until no blocking findings remain.
The curator proposes and stages only locator-backed rows; the auditor appends findings but cannot edit canonical rows.
In deployment, a human acceptance step gates final sign-off; this paper evaluates the LLM-curator + LLM-auditor sub-pipeline only.

\noindent\textbf{Source-gated acceptance and audit checks.}
We say that an accepted row has a non-parametric witness with respect to $M$ if every emitted row $R$ has an evidence locator $E(R)$ such that the locator content supports $R$ under an extractor $f$ that operates only on that locator content.
Stage--Audit operationalizes this through a source-citation gate: proposed rows without locators are rejected before audit.
The gate does not certify truth. It only shifts the failure mode: a row is wrong only if the extractor misread the cited locator.
The artifact contract $A=(Q,K,C,P,S,E)$ records the user query, primary-key schema, columns, scope statement, source set, and evidence map.
Audit checks span row evidence, key integrity, schema conformance, cardinality, source-role coverage, claim type, normalization, scope match, and temporal knowability (checks: App.~\ref{app:checks}; severity: App.~\ref{app:severity}; proof sketch: App.~\ref{app:proof}).
\emph{Factual} findings target row evidence; \emph{structural} findings target $K{,}C$; \emph{scope-related} findings target $P{,}S$.
Figure~\ref{fig:system} shows two scope/cardinality findings the citation gate alone would miss.

\subsection{Seed2Frontier Discovery Procedure}
\label{sec:method:discovery}

Seed2Frontier takes a user question and a seed Wiki page, then asks for the complement pages needed to assemble a structured table.
Stage--Audit bounds search through a logged frontier discipline: every page in $S$ records the operator that admitted it (seed table, sibling list, redirect, entity page, optional Wikidata count or rank check; template in Appendix~\ref{app:sparql}).
Audit treats missing required operators, unexplained pages, and mismatched source roles as scope failures.
Full pseudocode appears in Appendix~\ref{app:discovery}.
The procedure is a source-discovery policy, not a new retrieval method; existing planners can attempt the same task, and our experiment isolates the value of the governance overlay.

\section{Anchor Experiment: Seed2Frontier Discovery}
\label{sec:experiment}

\textbf{Curated evaluation set.} We curate 51 Seed2Frontier instances spanning 15 top-level domains (geography, culture, science, politics, sports, business, technology, and others).
35 instances are independently authored questions; 16 are programmatic derivatives produced by filter, top-K, or year-range modification of multi-page parents (Appendix~\ref{app:benchmark}).
Each instance has a Wikipedia seed page, labelled complement pages (45 of 51, mean 10.1 per instance, range 1--35), and a primary-key ground-truth table.

\textbf{Configurations.} Four conditions evaluated with the same model.
\emph{Memory-only}: one prompt for parametric-memory rows, no source frontier produced.
\emph{Seed-outlink}: deterministic Wikipedia-API enumeration of all outlinks of the seed; no LLM, no rows.
\emph{Vanilla LLM curator}: a single LLM prompt asking for complement pages and rows, no source-citation gate, no audit (a proxy for an unguarded LLM-based retrieval planner).
\emph{Stage--Audit-governed}: curator prompt with the Wikipedia discovery hint and source-citation requirement, followed by an auditor prompt that appends findings under the 12-check taxonomy.

\textbf{Setup.} We run the experiment with \texttt{gpt-5.4} at $T{=}0$.\footnote{Identifiers \texttt{gpt-5.4} and \texttt{Llama 3.3 70B} are API release tags, quoted verbatim from the API response.}
We manually label the seed Wikipedia URL for each instance as the most natural starting page among the candidate URLs.
Two protocol steps are ablated: human acceptance (Figure~\ref{fig:system}, gray dashed) and any repair-restage iteration. We run a single curator-and-auditor pass to match the single-pass budget the vanilla and memory-only baselines receive (loop ablation in Appendix~\ref{app:loop}).
Following WikiTabGen-style scoring (Appendix~\ref{app:scoring}), we report recall, precision, and F1 over two axes: predicted Wikipedia pages against the labelled complement set, and emitted rows against the labelled primary keys.

\begin{table}[t]
  \centering
  \small
  \caption{Stage--Audit on the 51-instance Seed2Frontier evaluation set with gpt-5.4 at $T{=}0$. Frontier metrics (primary axis) on the 45 instances with complement-page labels; PK metrics (secondary axis, see body) on all 51. Closed-loop scoring: when the auditor returns no accepted rows or pages (8/51 for rows, 4/51 for pages), Stage--Audit emits 0 for that instance rather than passing curator output through.}
  \label{tab:anchor}
  \resizebox{\linewidth}{!}{%
  \begin{tabular}{lccccccc}
    \toprule
    & \multicolumn{4}{c}{Frontier (page set)} & \multicolumn{3}{c}{Primary key (row set)} \\
    \cmidrule(lr){2-5} \cmidrule(lr){6-8}
    Config & Recall & Prec. & F1 & Size & Recall & Prec. & F1 \\
    \midrule
    Memory-only & n/a & n/a & n/a & 0.0 & \textbf{0.644} & \textbf{0.645} & \textbf{0.619} \\
    Seed-outlink & \textbf{0.580} & 0.020 & 0.034 & 612.3 & n/a & n/a & n/a \\
    Vanilla LLM curator & 0.384 & 0.356 & 0.334 & 15.1 & 0.598 & 0.611 & 0.588 \\
    Stage--Audit-governed & 0.491 & \textbf{0.505} & \textbf{0.451} & 11.4 & 0.544 & 0.599 & 0.534 \\
    \bottomrule
  \end{tabular}}
\end{table}

\textbf{Frontier comparison.}
Stage--Audit improves source-frontier precision (0.505 vs.\ 0.356), F1 (0.451 vs.\ 0.334), and complement recall (0.491 vs.\ 0.384) over the vanilla LLM curator on the 45 instances with complement labels.
We compute paired-bootstrap 95\% CIs by resampling the 45 per-instance Stage--Audit$-$vanilla differences with replacement (5000 draws) and taking the central 95\% range of the resulting means: $[+0.05,+0.25]$ for precision (two-sided sign test $p{=}0.006$), $[+0.03,+0.22]$ for F1 ($p{=}0.029$), and $[+0.02,+0.21]$ for recall.
The precision and F1 intervals lie entirely above zero, so the gains are robust to instance resampling.
The seed-outlink-vs-vanilla gap (precision 0.020 vs.\ 0.356) shows that LLM-based discovery does most of the precision lift over deterministic outlink enumeration; Stage--Audit then adds the source-citation gate and the audit step, raising precision a further +0.149 absolute (+42\% relative) while producing per-row locator output the unguarded curator does not.

\textbf{PK rows (secondary axis).}
By construction Stage--Audit emits only source-grounded rows, so unguarded baselines that emit every remembered row rank higher on PK overlap by design (memory-only $0.619$ $>$ vanilla $0.588$ $>$ Stage--Audit $0.534$); the primary axis is frontier quality with per-row source traceability, and a memory-only table cannot satisfy a source-grounded consumer.

\textbf{Where the policy helps.}
Stage--Audit's frontier-F1 gain over vanilla is largest on small-frontier instances (1--3 complement pages, $+0.25$, $n{=}17$); the gain shrinks on larger frontiers (4--9 pages, $+0.07$, $n{=}10$; 10+ pages, $+0.02$, $n{=}18$).
The largest per-domain gains are in \emph{culture} ($+0.32$, $n{=}7$), \emph{sports} ($+0.25$, $n{=}4$), and \emph{science} ($+0.18$, $n{=}5$); $n{\leq}3$ buckets are deferred to the supplement.

\textbf{Audit-finding density.} 182 findings across 51 instances (mean 3.6, range 1--5; issue-type$\times$severity in Figure~\ref{fig:findings}, Appendix~\ref{app:findings-figure}); the auditor returned no accepted rows on 8/51 instances and no accepted complement pages on 4/51, contributing 0 to the Stage--Audit means under closed-loop scoring.
Stage--Audit's frontier averages 11.4 pages per instance, $54\times$ smaller than seed-outlink's 612.3.

\textbf{Cross-model.} Repeating the four-configuration evaluation on Llama 3.3 70B (open-source; Appendix~\ref{app:cross-model}) yields a small directional gain ($+0.008$ frontier F1 vs.\ vanilla, $+0.012$ precision, $+0.033$ recall); the policy contribution is most pronounced on stronger instruction-following models.

\section{Conclusion}
\label{sec:conclusion}

Stage--Audit governs Seed2Frontier with disjoint curator-auditor write rights, a row-level source-citation gate, and a 12-check audit taxonomy.
On 51 instances using gpt-5.4 it improves source-frontier precision over an unguarded LLM curator from 0.356 to 0.505 (paired 95\% CI $[+0.05,+0.25]$, $p{=}0.006$) and F1 from 0.334 to 0.451.
The present evaluation ablates the human-acceptance step and the repair-restage loop, and covers one closed-source and one open-source model.
Within those scope limits, Stage--Audit serves as a source-grounding integrity overlay for cross-Wiki table construction.

\bibliographystyle{plainnat}
\bibliography{refs}

\appendix
\section{Seed2Frontier Discovery Algorithm}
\label{app:discovery}

\begin{algorithm}[H]
\caption{Seed2Frontier discovery procedure (curator's propose step in Figure~\ref{fig:system})}
\begin{algorithmic}[1]
\Require natural-language table query $Q$, seed Wiki page $s$
\Ensure candidate source set $S$
\State $P \gets \textsc{InferScope}(Q)$ \Comment{partition/scope statement extracted from $Q$ (e.g., ``by continent'' $\Rightarrow$ continents enumeration)}
\State $S \gets \{s\}$ with source role and retrieval metadata
\If{a single page covers all partitions in $P$}
  \State \Return $S$
\EndIf
\State expand through sibling list pages, redirects, and list/table links
\For{each under-covered partition or tail entity in $P$}
  \State fetch the relevant entity page and add supporting pages to $S$
\EndFor
\If{$Q$ implies a count, rank, or partition-size constraint}
  \State run a Wikidata SPARQL count or rank sanity check
\EndIf
\State log each URL, locator, retrieval time, and content hash for audit
\State \Return $S$
\end{algorithmic}
\end{algorithm}

\section{Wikidata Sanity-Check Template}
\label{app:sparql}

When a user question implies a count, rank, membership, or partition-size constraint, Wikidata can provide a weak sanity check on the source frontier.
The query is not the row witness: accepted rows still require source locators in $E$.
In practice, the curator records the property and class choices, the query time, and the returned count or rank signal so an auditor can replay whether the frontier is plausibly complete.

\begin{verbatim}
# Q_CLASS, P_FILTER, Q_FILTER are placeholders for instance-specific
# Wikidata classes and properties; substitute before issuing the query.
PREFIX wd: <http://www.wikidata.org/entity/>
PREFIX wdt: <http://www.wikidata.org/prop/direct/>
SELECT (COUNT(DISTINCT ?entity) AS ?n) WHERE {
  ?entity wdt:P31/wdt:P279* wd:Q_CLASS .
  ?entity wdt:P_FILTER wd:Q_FILTER .
}
\end{verbatim}

\section{Proof Sketch for the Non-Parametric Witness Claim}
\label{app:proof}

The source-citation gate gives a simple invariant over staged rows.
Let $R_A$ be the rows in artifact $A$ and let $\mathrm{Block}(A)$ be its unresolved blocking findings.
For row $r$, define the operational acceptance predicate:
\[
\mathrm{RowOK}(r,A) \equiv \exists e \in E(r) : e \in S \wedge \mathrm{supports}_f(e,r).
\]
The acceptance condition is then:
\[
\mathrm{Accept}(A) \equiv \forall r \in R_A,\ \mathrm{RowOK}(r,A) \wedge \mathrm{Block}(A)=\emptyset.
\]
Here $\mathrm{supports}_f(e,r)$ means that extractor $f$ reads locator content $e$, rather than model memory, and recovers support for the row.
For any accepted row $r$, the staging rule requires such an evidence locator in the logged source set $S$.
The audit rule then permits validation only from the locator content, not from the model's remembered facts.
By induction over staged row updates, every row that passes the gate has such a witness, modulo extractor faithfulness and locator drift.

\section{Curated Evaluation Set}
\label{app:benchmark}

We curate 51 Seed2Frontier instances spanning 15 top-level domains, with the largest concentrations in geography (13 instances), culture (8), science (6), politics (4), sports (4), and business (3); the remaining nine domains (technology, religion, food, education, environment, health, history, law, society) contribute one to three instances each.
Each instance is paired with a Wikipedia seed page, the labelled set of cited Wikipedia URLs (mean 10.1 pages per instance, median 6, range 1--35), and a primary-key ground-truth table (mean 68.2 rows per instance, median 24, range 3--526).
We manually label the seed page for each instance as the most natural starting Wikipedia page for the query; the remaining labelled Wikipedia URLs form the complement set.

\section{Scoring Notes}
\label{app:scoring}

Complement-page recall compares predicted source pages to the labelled complement set.
Source-frontier precision and F1 use the same page set after Wikipedia-title normalization.
Cardinality error is the absolute difference between emitted row count and labelled row count.
Table-extraction F1 is computed over normalized primary-key overlap when a configuration emits rows.

\section{Severity Taxonomy Reference}
\label{app:severity}

Issue types are \emph{factual} (a source-supported value is wrong or unsupported), \emph{structural} (schema, primary key, type, duplicate, or artifact-contract problem), and \emph{scope-related} (row-universe, partition, rank, cardinality, or exhaustiveness problem).
Severity levels are \emph{blocking}, \emph{hygiene}, and \emph{suggestion}.
The crossing is deliberate: a scope-related blocker can make a table unusable even if every emitted row is factually correct, while a structural hygiene issue can be repairable without changing the row set.

\section{Cross-Model Generalization}
\label{app:cross-model}

We pick one closed-source model (gpt-5.4) and one open-source model (Llama 3.3 70B) as the two model families $M$, to test whether the policy contribution generalizes across release types and training pipelines.

\begin{table}[ht]
  \centering
  \small
  \caption{Vanilla LLM curator vs.\ Stage--Audit on the same 51-instance Seed2Frontier evaluation set with picked seeds, across one closed-source and one open-source model. Frontier metrics on the 45 instances with complement labels. $\Delta$ is Stage--Audit minus vanilla; sign-test $p$ on per-instance F1 deltas.}
  \label{tab:cross-model}
  \begin{tabular}{lccccccc}
    \toprule
    & \multicolumn{3}{c}{Vanilla LLM curator} & \multicolumn{3}{c}{Stage--Audit-governed} & \\
    \cmidrule(lr){2-4} \cmidrule(lr){5-7}
    Model & Rec. & Prec. & F1 & Rec. & Prec. & F1 & $\Delta$F1 (sign-$p$) \\
    \midrule
    gpt-5.4          & 0.384 & 0.356 & 0.334 & \textbf{0.491} & \textbf{0.505} & \textbf{0.451} & \textbf{$+0.118$ ($0.014$)} \\
    Llama 3.3 70B    & 0.329 & 0.344 & 0.311 & 0.362 & 0.356 & 0.319 & $+0.008$ ($0.21$) \\
    \bottomrule
  \end{tabular}
\end{table}

Stage--Audit's frontier-F1 improvement over the vanilla LLM curator is large and paired-significant on gpt-5.4 ($+0.118$, two-sided sign test $p{=}0.029$) and small but directionally consistent on Llama 3.3 70B ($+0.008$ F1, with positive deltas on all three frontier metrics).
The pattern is consistent with the policy contribution being most effective on top of strong-instruction-following models, where the curator emits citation-tagged rows the auditor can then filter cleanly.

\section{Severity Taxonomy in the 182 Auditor Findings}
\label{app:findings-figure}

\begin{figure}[ht]
  \centering
  \includegraphics[width=0.96\linewidth]{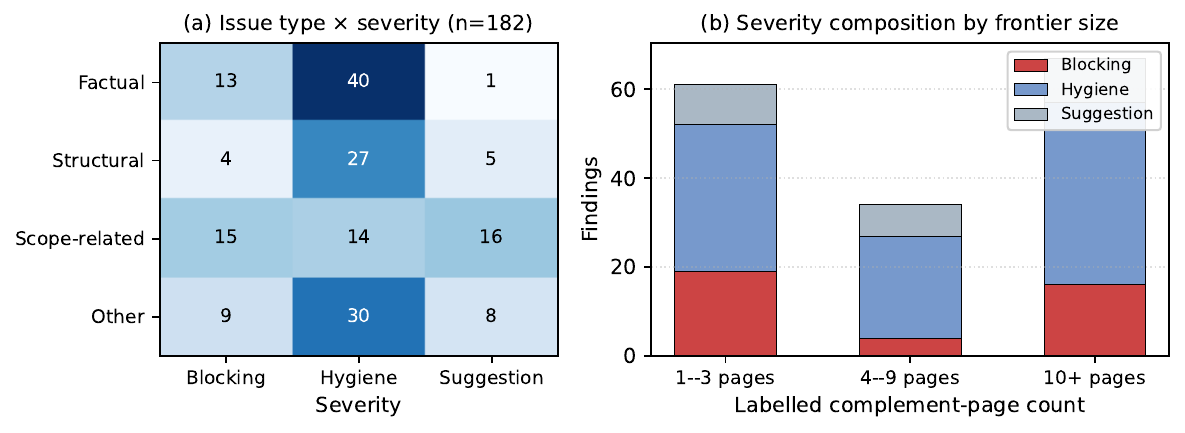}
  \caption{Severity taxonomy applied to the 182 Stage--Audit auditor findings on the 51-instance set with gpt-5.4. (a) Issue-type $\times$ severity heatmap: factual findings are dominated by the source-citation gate ($n{=}40$ hygiene, 13 blocking) confirming the gate fires on essentially every instance; scope-related findings are spread across blocking, hygiene, and suggestion, reflecting the harder-to-automate cardinality and partition checks. (b) Severity composition stratified by labelled complement-page count: every frontier-size bucket produces blocking findings, and the smallest-frontier bucket carries the highest blocking share, consistent with the Section~\ref{sec:experiment} finding that the gate provides the most lift on small-frontier instances. Issue-type assignments use a conservative keyword classifier over the auditor's free-text \texttt{check} labels; the small ``Other'' bucket reflects labels that did not match the keyword set.}
  \label{fig:findings}
\end{figure}

\section{Audit Check Enumeration}
\label{app:checks}

The 12-check audit taxonomy used in Section~\ref{sec:method}. Each check has a stable identifier; the auditor assigns one identifier per finding.

\noindent\textit{Factual.} \texttt{F1}\,row-evidence (locator content supports row); \texttt{F2}\,source-URL well-formed and reachable; \texttt{F3}\,locator format and section/table reference valid.

\noindent\textit{Structural.} \texttt{S1}\,primary-key uniqueness; \texttt{S2}\,primary-key non-null; \texttt{S3}\,column type conformance; \texttt{S4}\,column completeness against the artifact contract.

\noindent\textit{Scope-related.} \texttt{P1}\,cardinality match against query expectation; \texttt{P2}\,partition coverage; \texttt{P3}\,source-role coverage (each page tagged with its admitting operator); \texttt{P4}\,temporal knowability; \texttt{P5}\,normalization consistency across rows.

\section{Repair-Loop Ablation}
\label{app:loop}

The body experiment caps Stage--Audit at a single curator-and-auditor pass to match the per-instance call budget of the unguarded baselines.
To check whether additional repair-restage iterations would change the picture, we run gpt-5.4 with $k{=}0,\dots,5$ repair iterations on a 15-instance domain-stratified subset (one instance per top-level domain).
Each iteration after $k{=}0$ feeds the auditor's findings back to the curator with a restage prompt and re-audits the new output; an instance converges if the auditor returns no blocking findings.

\begin{table}[ht]
  \centering
  \small
  \caption{Stage--Audit performance vs.\ repair-loop count $k$ on a 15-instance domain-stratified subset (gpt-5.4 at $T{=}0$); $k{=}0$ is the body experiment. PK-F1 gains $+0.02$ at $k{=}1$ and plateaus.}
  \label{tab:loop}
  \begin{tabular}{cccccc}
    \toprule
    $k$ & Comp.\ recall & Frontier prec. & Frontier F1 & PK F1 & Frontier size \\
    \midrule
    0 & 0.403 & 0.404 & 0.375 & 0.544 & 10.1 \\
    1 & 0.403 & 0.401 & 0.373 & \textbf{0.565} & 10.5 \\
    2 & 0.403 & 0.401 & 0.373 & 0.560 & 11.9 \\
    3 & 0.403 & 0.401 & 0.373 & 0.560 & 10.8 \\
    \bottomrule
  \end{tabular}
\end{table}

Stage--Audit's frontier metrics are stable through $k{=}3$ and show no improvement past the initial pass; PK-F1 receives a small one-shot lift at $k{=}1$ and then plateaus.
Productive iteration would require either external retrieval or auditor signals naming candidate replacement pages, both deferred to future work.

\end{document}